\documentclass[11pt]{article}

\ifdefined\PsyCIDRAReview
  \usepackage[review]{acl}
\else
  \usepackage[preprint]{acl}
\fi
\usepackage{times}

\usepackage{latexsym}
\usepackage[T1]{fontenc}
\usepackage[utf8]{inputenc}
\usepackage{microtype}
\usepackage{graphicx}
\usepackage{booktabs}
\usepackage{tabularx}
\usepackage{amsmath}
\usepackage{xcolor}
\graphicspath{{figures/}}
\title{PsyCIDRA: A Dual-Agent Framework for Psychiatric Interviewing and Diagnostic Reasoning}

\ifdefined\PsyCIDRAReview
\author{Anonymous ACL Submission}
\hypersetup{pdfauthor={Anonymous}}
\else
\author{
  \bfseries Milad Mohammadi$^{1}$ \quad
  Fatemeh Akrami Shamsabadi$^{2}$ \quad
  Zahra Mohseni$^{2}$ \\[2pt]
  \bfseries Amirhossein Safdarian$^{1}$ \quad
  Malekfarhad Malek$^{3}$ \quad
  Hadi Moradi$^{1}$ \quad
  Hesham Faili$^{1}$ \\[6pt]
  \normalfont $^{1}$School of Electrical and Computer Engineering, University of Tehran \\
  \normalfont $^{2}$Faculty of Psychology and Educational Sciences, University of Tehran \\
  \normalfont $^{3}$School of Medicine, Yazd (Shahid Sadoghi) University of Medical Sciences \\[6pt]
  \normalfont\texttt{\{miladmohammadi,akrami.f,z.mohseni,a.safdarian,moradih,hfaili\}@ut.ac.ir} \\[2pt]
  \normalfont\texttt{mf.malek@ssu.ac.ir}
}
\hypersetup{pdfauthor={Milad Mohammadi, Fatemeh Akrami Shamsabadi, Zahra Mohseni, Amirhossein Safdarian, Malekfarhad Malek, Hadi Moradi, Hesham Faili}}
\fi
\hypersetup{pdftitle={PsyCIDRA: A Dual-Agent Framework for Psychiatric Interviewing and Diagnostic Reasoning}}

\begin{document}
\raggedbottom
\maketitle

\begin{abstract}
Large language models show promise in clinical reasoning, but psychiatric interviewing requires guiding an evolving conversation. Their ability to carry out this interactive assessment remains less studied. We present PsyCIDRA, a dual-agent framework linking free-form psychiatric interviewing with diagnostic reasoning for expert review. Its interviewer agent uses tools to maintain working notes, load expert-written skills, and retrieve ICD-11 references to guide inquiry. Its diagnostic reasoning agent then receives the completed interview transcript and reports hypotheses alongside supporting, conflicting, and missing evidence, withholding a final hypothesis when none is sufficiently supported. Using patient profiles generated with PsyCPG, we first evaluate PsyCIDRA in simulation. Across four models on 53 evaluation cases, it achieves higher diagnostic agreement than direct prompting. On 81 held-out simulated cases, rank-1 accuracy is 60.5\% versus 51.9\%. In a blinded study of 101 human participants in separate arms, PsyCIDRA agrees with psychologists on whether to propose a diagnostic hypothesis in 79.6\% of cases, compared with 65.4\% for direct prompting. Together, these findings support the potential of LLM agents to assist psychiatric assessment through free-form dialogue. By examining diagnostic reasoning, interview quality, and safety together, this study contributes to understanding the capabilities and limitations of psychiatric interview agents.
\end{abstract}

\section{Introduction}
\label{sec:introduction}

Mental disorders affect more than one billion people worldwide, yet access to assessment and care remains limited \citep{who-2025-mental-health}. Even for major depressive disorder, only an estimated 9.1\% of affected people received minimally adequate treatment globally in 2021 \citep{santomauro-2024-treatment}. Timely recognition and diagnosis are important steps toward appropriate care \citep{who-2024-icd11-cddr}. Yet identifying a mental disorder requires more than recognizing a symptom: disturbances in cognition, emotion regulation, or behavior must be understood in relation to their duration, context, and impact on daily functioning.

To support diagnostic assessment, clinicians can draw on mental-health questionnaires and structured or semistructured interview instruments. Questionnaires offer standardized symptom screening, but most target individual disorders and require follow-up diagnostic assessment \citep{zimmerman-2024-questionnaires}. Structured interviews such as the MINI \citep{sheehan-1998-mini} and semistructured interviews such as the SCID \citep{first-2016-scid5-cv} guide diagnostic assessment through predefined modules, while requiring time and expertise. Branching questions and, in semistructured formats, additional clinical probing provide flexibility within this framework. More open-ended clinical interviewing requires clinicians to shape the inquiry around the person's account and emerging diagnostic possibilities \citep{holmes-2011-hypothesis-driven}, clarifying ambiguous descriptions and deciding which questions to pursue as the conversation unfolds. Large language models (LLMs) offer a promising basis for complementary tools that support this process by gathering and connecting evidence through dialogue for professional review.

\begin{figure*}[!t]
  \centering
  \setlength{\abovecaptionskip}{6pt}
  \includegraphics[width=\textwidth,trim=0bp 3bp 0bp 0.6bp,clip]{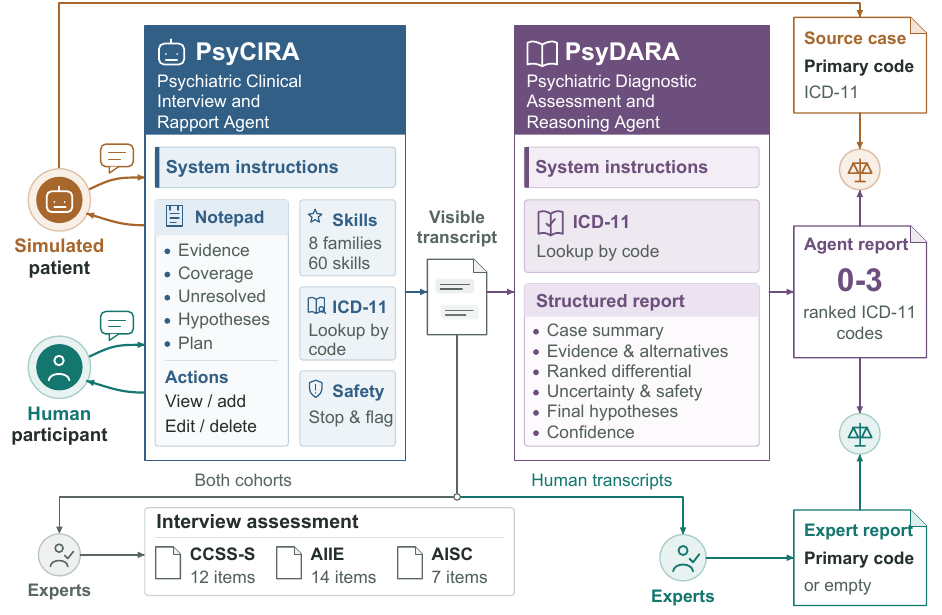}
  \caption{PsyCIDRA and its evaluation. Simulated patients use PsyCPG-generated profiles. PsyDARA analyzes only PsyCIRA's visible transcript. Case diagnoses and expert judgments provide simulation and human references, respectively; neither agent sees these references.}
  \label{fig:workflow}
\end{figure*}

Evidence for this promise comes from strong performance on medical question-answering benchmarks \citep{singhal-2025-medpalm2} and psychiatric assessment from existing clinical records and interview transcripts \citep{sun-2026-mentalseek-dx,rosenman-2024-questionnaire}. In these settings, however, the evidence is supplied to the model. Gathering it through dialogue is a less studied and more demanding task: the model must choose informative questions, resolve ambiguity across turns, and revise its inquiry as answers emerge. Evaluations of clinical agents reveal difficulties with incomplete information gathering and premature conclusions despite strong medical knowledge \citep{lievin-2026-residencyrl}. Psychiatric interviewing systems have begun to address this challenge \citep{bi-2025-magi,wu-2026-wisemind}, but evidence remains limited for free-form interviewing across diverse presentations with expert evaluation of both the interaction and its diagnostic conclusions. This calls for studying how an agent elicits evidence, alongside what it can infer from the resulting conversation.

We investigate this task through \textbf{PsyCIDRA}, a dual-agent system designed to support expert assessment. PsyCIRA determines question content and sequence from the unfolding conversation, guided by clinical instructions and references without a prescribed questionnaire or branching interview protocol. PsyDARA interprets only the resulting transcript, linking ranked diagnostic hypotheses to supporting, conflicting, and missing evidence. It can withhold a final hypothesis when evidence is insufficient or the reported presentation does not meet diagnostic criteria. Reports are intended for qualified experts to inspect and revise, and are not delivered to participants.

Our evaluation separates interviewing from diagnostic reasoning and combines controlled simulation with expert assessment of human interviews. We make three contributions:

\noindent (1) \textbf{PsyCPG: psychiatric patient simulation.} We introduce PsyCPG, a generation-and-validation pipeline that converts clinical cases into structured patient profiles for psychiatric patient simulation. We produce 212 quality-checked profiles and show higher expert-rated simulation quality than role-play directly from clinical cases.

\begin{samepage}
\noindent (2) \textbf{PsyCIDRA and simulated experiments.} We develop PsyCIDRA and evaluate its interviewing and diagnostic components across four models and multiple configurations. Experiments on a 53-case evaluation cohort and a separate 81-case held-out test set show higher diagnostic agreement than direct prompting of the same underlying models.
\par\end{samepage}

\noindent (3) \textbf{Human feasibility study.} We conduct a study with 101 participants in which psychologists review interview quality and diagnostic hypotheses. Compared with direct prompting, PsyCIDRA more often agrees with experts on whether an interview supports a diagnostic hypothesis.

\ifdefined\PsyCIDRAReview
All code, simulation profiles, expert reviews, and results are provided for review and will be publicly released upon publication.
\else
All code, simulated patient profiles, expert reviews of simulations, and experimental results are publicly available at {\urlstyle{same}\url{https://github.com/PsyCIDRA/PsyCIDRA}}.
\fi
Human-participant chat data will not be publicly released for privacy reasons.

\section{Related Work}
\label{sec:related-work}

AI-assisted psychiatric diagnosis predates LLMs \citep{dwyer-2018-machine-learning,shatte-2019-machine-learning,abd-alrazaq-2022-ai-diagnosis}. LLM-based studies predict mental-health conditions from social-media posts \citep{yang-2023-interpretable}, estimate symptom scores from recorded interviews \citep{rosenman-2024-questionnaire}, and examine diagnostic reasoning from clinical records \citep{sun-2026-mentalseek-dx}. These approaches analyze supplied evidence; interviewing systems must also decide what information to elicit.

Conversational assessment varies in both interview structure and diagnostic scope. Depression-focused systems select from predefined symptoms and conversational strategies \citep{cao-2025-stigma} or learn to probe the eight PHQ-8 symptom domains \citep{bi-2026-s4}. WiseMind follows DSM-5 decision paths for depressive, bipolar, and anxiety presentations \citep{wu-2026-wisemind}. MAGI allows rephrasing and clarification within the MINI interview tree; its human evaluation covers depression, generalized and social anxiety, and suicide risk, using psychologists' annotations \citep{bi-2025-magi}. These studies demonstrate responsive dialogue and assess its performance within a limited set of clinical targets.

Broader diagnostic coverage has also been explored, although the evidence differs across evaluation settings. LingxiDiagBench includes free-form consultation across 12 ICD-10 categories, evaluated through simulation and record-based role-play \citep{xu-2026-lingxidiagbench}. TalkToAlba assesses nine disorders in 303 human participants, moving from open exploration to structured DSM-5 confirmation \citep{sikstrom-2025-talktoalba}. Its diagnostic reference consists of participants' reports of prior clinician diagnoses, rather than expert assessment of the current interview.

Evidence remains limited for the combination of free-form clinical interviewing, coverage across diverse disorder families beyond a small set of target conditions, and expert assessment of both interview quality and transcript-supported diagnostic conclusions. Our study examines this combination through simulation and interviews with human participants.

\ifdefined\PsyCIDRAReview
  \newpage
\fi
\section{Psychiatric Patient Simulation}
\label{sec:simulation}

We designed and evaluated a system that simulates psychiatric patients from clinical cases and their diagnosis codes, providing a controlled environment for assessing clinical interviewing and diagnostic reasoning. Prior work has used manually authored patient profiles \citep{wang-2024-patient-psi} or focused on restricted diagnostic settings \citep{cao-2025-stigma,lee-2025-psyche}. Our evaluation required cohorts spanning a broader range of disorders, with varied presentations and a known clinical reference against which to assess the information elicited during an interview.

\begin{figure*}[t]
\def\simulationdisplayheight{142pt}
\begin{minipage}[t]{\columnwidth}
\vspace{0pt}
\begin{minipage}[c][\simulationdisplayheight][c]{\linewidth}
\centering
\includegraphics[width=\linewidth,height=\simulationdisplayheight,keepaspectratio]{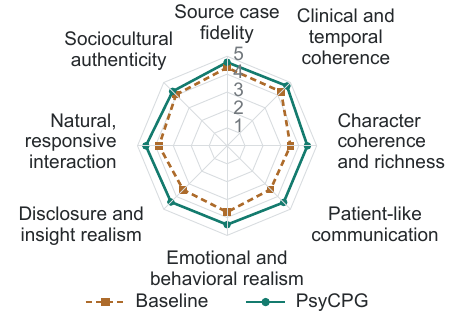}
\end{minipage}
\caption{Mean MH-SPQF ratings by dimension (1--5; higher is better).}
\label{fig:simulator-quality}
\end{minipage}\hfill
\begin{minipage}[t]{\columnwidth}
\vspace{0pt}
\begin{minipage}[c][\simulationdisplayheight][c]{\linewidth}
\centering
\small
\renewcommand{\arraystretch}{1.25}
\setlength{\tabcolsep}{3pt}
\begin{tabular}{@{}lrr@{}}
\toprule
Measure & Baseline & PsyCPG \\
\midrule
Mean quality score & 3.83/5 & 4.50/5 \\
Overall authenticity & 3.47/5 & 4.50/5 \\
Intended diagnosis match & 24/30 (80\%) & 29/30 (97\%) \\
\midrule
\multicolumn{3}{@{}l@{}}{\textit{Error flags (less is better)}} \\
Any error flag & 16/30 & 4/30 \\
Excessive disclosure & 11/30 & 3/30 \\
Artificial expression & 12/30 & 2/30 \\
Unsupported facts & 2/30 & 1/30 \\
\bottomrule
\end{tabular}
\end{minipage}
\captionof{table}{Expert evaluation of raw-case simulation (Baseline) and simulation using PsyCPG-generated profiles on 30 paired cases.}
\label{tab:simulator}

\end{minipage}
\end{figure*}

\subsection{PsyCPG}
PsyCPG (Psychiatric Case-based Profile Generation) aims to simulate patients with rich, varied personal and sociocultural characteristics while prioritizing clinical coherence and fidelity to the source case. We used MentalDx Bench, which provides deidentified electronic health records with psychiatrist-assigned ICD-11 diagnoses \citep{sun-2026-mentalseek-dx}. Before profile generation, we partitioned 320 selected cases into three disjoint cohorts: simulator validation, evaluation, and held-out test. Quality filtering retained 78, 53, and 81 profiles; collectively, they cover 63 diagnostic codes across 16 groups.

PsyCPG converts each source case and its ICD-11 reference into a reusable patient representation through generation, validation, and targeted revision. An LLM generates a cohesive character narrative conditioned on the case, diagnostic requirements, sampled Big Five traits, and sociocultural attributes selected from a predefined catalogue. These additional inputs specify variation in perspective and communication, while generation instructions preserve explicit source facts and restrict diagnostic enrichment to subjective experiences and functional impact. A separate LLM validation call checks source fidelity, diagnostic consistency, and unsupported high-stakes content, and extracts a typed fact ledger. Fact provenance distinguishes source evidence, diagnosis-informed elaboration, and nonclinical character development, with schema constraints limiting which fact types each may contribute. Validation warnings trigger one targeted revision followed by revalidation; profiles with material errors are excluded.

Each accepted profile is rendered into the system prompt of an LLM patient agent, which generates each response from the profile and accumulated dialogue history. The narrative, demographics, and personality traits guide linguistic style and interpersonal behavior, while the fact ledger supplies explicit content constraints. Each ledger entry records a fact, its semantic type, timeframe, and disclosure level. These levels define a prompt-based disclosure policy: information may be volunteered, introduced when its topic becomes relevant, reserved for direct questions, or initially withheld when sensitive. The policy separates the simulator's available knowledge from what the patient is instructed to reveal at a given turn. Thus, dialogue context governs which grounded facts the agent should express.

Safety settings keep risk-related behavior consistent with the case, while configurable edge behaviors, such as topic avoidance, enable controlled tests of challenging interactions. Together, these components target excessive disclosure, clinician-like language, and factual inconsistency documented in prior work \citep{wang-2024-patient-psi,kyung-2025-patientsim}. Reference diagnosis labels and codes are withheld during role-play, although diagnosis-informed enrichment can still introduce cues.

\subsection{Evaluation}
We compared patient simulation using PsyCPG-generated profiles with a baseline that received the same raw clinical case and a role-play instruction. Two psychology experts directly interviewed both conditions on 30 paired cases from the simulator-validation cohort, yielding 60 interviews. Both conditions used GPT-5.6 Terra with the same reasoning setting and a 16-exchange limit. Each expert conducted 15 interviews per condition and saw only one version of each case; the other expert interviewed its paired counterpart. During interviewing, experts were blinded to condition, source case, and reference diagnosis. They recorded diagnostic hypotheses before the source material was revealed for quality assessment.

With psychology experts, we developed the \emph{Mental Health Simulated Patient Quality Form} (MH-SPQF). It combines eight ratings of clinical fidelity and interaction realism, a separate overall authenticity rating (all 1--5), and five binary error flags. The mean quality score averages the available dimension ratings within each interview, then across interviews. Overall authenticity captures the expert's holistic judgment of resemblance to a real patient. Error flags identify specific failures, including unsupported clinical facts, excessive disclosure, and artificial expression. This study-specific form has not undergone psychometric validation.

Experts rated patients simulated from PsyCPG-generated profiles higher across all eight dimensions (Figure~\ref{fig:simulator-quality}). The mean quality score increased from 3.83 to 4.50, and overall authenticity from 3.47 to 4.50 (Table~\ref{tab:simulator}). The largest gains concerned patient-like expression and realistic disclosure and insight. At least one error was flagged in 4/30 interviews with PsyCPG-generated patients, compared with 16/30 baseline interviews. Excessive disclosure fell from 11 to 3 interviews and artificial expression from 12 to 2. This pattern is consistent with the intended role of the behavioral and disclosure controls, although the study does not isolate their individual contributions.

The experts' first diagnosis matched the intended ICD-11 code in 29/30 interviews with PsyCPG-generated patients, compared with 24/30 baseline interviews. Five cases improved and none worsened. Together with higher ratings for disclosure and insight realism and fewer excessive-disclosure flags, this pattern suggests that PsyCPG-generated profiles better align simulated behavior with the intended diagnosis while keeping information sharing appropriately paced and selective. Although this study of 30 paired cases does not establish equivalence to real patients, the gains in clinical fidelity and interaction realism support using the profiles for controlled validation of interviewing and diagnostic systems.

\section{PsyCIDRA}
\label{sec:method}

PsyCIDRA (Psychiatric Clinical Interview and Diagnostic Reasoning Agents) couples two LLM agents with tool access and complementary roles (Figure~\ref{fig:workflow}). PsyCIRA (Psychiatric Clinical Interview and Rapport Agent) conducts the interview, and PsyDARA (Psychiatric Diagnostic Assessment and Reasoning Agent) converts the completed transcript into a report for expert review. NLP researchers and psychologists jointly designed the system's prompts, tools, skill catalogue, and report schema.

Several related mental-health systems use workflow-based architectures to coordinate specialized components. MAGI separates navigation, questioning, judgment, and diagnosis within a MINI-guided workflow \citep{bi-2025-magi}, while the cognitive layer of \citet{rollwage-2026-cognitive-layer} combines safety screening, clinical reasoning, retrieval, and response moderation for psychotherapy. Agents offer a different approach to orchestration by interleaving reasoning, actions, and feedback \citep{yao-2023-react}. We investigate an agent design for interviewing and diagnostic assessment, giving each model control over when to consult its available resources and how to continue from their outputs. The aim is flexible action selection with fewer unnecessary calls and potentially lower latency.

\subsection{PsyCIRA: clinical interviewing}
PsyCIRA's system instructions provide general guidance for managing the interview session, maintaining rapport, and using its available tools.

Accumulating interaction histories can obscure the information needed for an agent's next action; explicit working-memory management has improved performance on long-horizon tasks \citep{hu-2025-hiagent}. We address this problem through a private \textbf{interview notepad}, inspired by note-taking during clinical interviews. It maintains five categories: evidence, covered topics, unresolved issues, provisional hypotheses, and a short-term plan. Read and update tools let the agent inspect the current notes and add, edit, or delete entries using stable identifiers. Instructions favor compact notes that preserve timing, negation, source, and uncertainty, with updates when evidence or priorities change. The plan identifies the next one to three useful questions, making follow-up priorities explicit across turns.

Reusable skills provide LLM agents with procedural guidance for recurring situations and can improve task performance \citep{wang-2025-awm,mi-2026-skill-pro}. Our expert-written \textbf{skill catalogue} contains 60 skills across eight categories. Whereas the system instructions define general interviewing principles, skills provide situational guidance. For example, a skill for guarded responses recommends shorter questions and respect for refusal.

ICD-11 provides disorder-specific clinical descriptions and diagnostic requirements, including essential features and distinctions from related conditions \citep{who-2026-icd11-mms}. PsyCIRA's \textbf{ICD-11 reference tools} load the relevant material to guide follow-up questions and clarify diagnostic boundaries.

For urgent concerns such as self-harm, harm to others, or acute medical danger, a \textbf{safety tool} limits further LLM interaction. Invoking it stops the interview and refers the participant for human review.

\subsection{PsyDARA: diagnostic reasoning}
PsyDARA receives only the completed transcript and operates in a tool-use loop with an ICD-11 disorder lookup tool. Its instructions guide it to formulate the presenting problem, identify plausible candidates, and consult their reference entries before assessing diagnostic fit. Independent lookups can run in parallel. The agent must distinguish what the interview establishes from what remains uncertain.

Generating intermediate reasoning before an answer allows autoregressive LLMs to condition later predictions on that preceding text \citep{merrill-2024-expressive}. We therefore design PsyDARA's \textbf{structured report} to place evidence and comparative reasoning before the final hypotheses. This ordering is intended both to guide generation and to provide an explicit explanation through stated evidence, alternatives, and uncertainties that experts can inspect.

The schema begins with data sufficiency and a compact problem representation, followed by evidence marked as present, denied, unclear, or unassessed. It considers alternative explanations and ranks a differential using supporting, conflicting, and missing evidence, then synthesizes the ranking and uncertainties and provides a separate safety assessment. The final field contains zero to three ranked hypotheses with ICD-11 base codes and qualitative confidence (high, moderate, or low). It remains empty when no disorder is meaningfully supported; incomplete evidence can lower confidence in an otherwise useful hypothesis.

\section{Experimental Design}
\label{sec:evaluation}

We evaluate PsyCIDRA first with simulated patients and then in a human-participant feasibility study. Interviews were conducted in Persian, and diagnostic reports were generated in English. Direct-prompting comparisons assess what the agent harness adds to the underlying LLMs through diagnostic agreement, interview quality, and safety.

The \emph{Direct} baseline prompts the interviewing LLM with task instructions and conversation history, and the diagnosis LLM with the resulting transcript to obtain ranked hypotheses. Within comparisons, conditions use the same backbone, reasoning setting, and 16-exchange budget. The contrast includes differences in prompts, tools, and computation.

\subsection{Simulated-patient experiments}
The evaluation cohort of 53 PsyCPG-generated profiles (Section~\ref{sec:simulation}) supports development and all four combinations of Direct/PsyCIRA interviewing and Direct/PsyDARA diagnosis. Each interviewer produces one transcript per case, reused by both diagnosers. Two psychologists assessed interview quality and safety on 46 Terra transcripts from 30 evaluation cases (23 per interviewer), blinded to interview mode and model-generated reports.

GPT-5.6 Terra with medium reasoning is the backbone model for interviewing, diagnosis, and patient simulation. Evaluation-cohort comparisons additionally use GPT-5.6 Luna without reasoning and DeepSeek V4 Flash/Pro with adapter defaults \citep{openai-gpt56-terra,openai-gpt56-luna,deepseek-v4-changelog}.

We then compare Direct with complete PsyCIDRA on the disjoint 81-profile held-out cohort. The held-out test and human study use Terra with configurations frozen after evaluation.

\subsection{Human-participant study}
We conducted a feasibility study with adult volunteers through a research web platform. Of 143 registrants, 113 had completed or safety-stopped interviews. Twelve interviews were excluded, including four early safety stops with fewer than six participant responses, leaving 101 for analysis (52 Direct, 49 PsyCIRA). Eligibility required at least six responses. Automated, balanced random allocation at registration assigned participants to the two interview modes (Appendix~\ref{app:human-attrition}); participants were blinded to interview mode.

Two psychologists independently assessed each transcript's interview quality and safety and proposed diagnoses, blinded to interview mode and model-generated reports. Disagreements in primary diagnostic codes or whether a diagnosis was warranted were resolved through discussion. Complete-system comparisons involve different participants. Both diagnosers also process all 101 transcripts, enabling comparison of diagnostic reasoning on shared evidence.

\begin{table*}[t]
  \centering
\begingroup
\centering
\fontsize{9}{10}\selectfont
\setlength{\tabcolsep}{3.2pt}%
\renewcommand{\arraystretch}{0.95}%
\begin{tabular*}{\textwidth}{@{\extracolsep{\fill}}llrrrrrrr@{}}
\toprule
\multicolumn{2}{c}{Configuration} & \multicolumn{4}{c}{Reference coverage by category (\%)} & \multicolumn{3}{c}{Overall ($n=53$)} \\
\cmidrule(lr){1-2}\cmidrule(lr){3-6}\cmidrule(l){7-9}
Interview & Diagnosis & \shortstack{Anxiety/fear\\$n=19$} & \shortstack{OC-related\\$n=15$} & \shortstack{Mood\\$n=10$} & \shortstack{Other\\$n=9$} & \shortstack{Coverage\\(\%)} & \shortstack{Rank 1\\(\%)} & MRR \\
\midrule
\multicolumn{9}{@{}l}{\textit{GPT-5.6 Terra}} \\
Direct & Direct & 84.2 & 86.7 & 80.0 & 55.6 & 79.2 & 73.6 & 0.764 \\
Direct & PsyDARA & 78.9 & 100.0 & 70.0 & 55.6 & 79.2 & 75.5 & 0.774 \\
PsyCIRA & Direct & 78.9 & 100.0 & 100.0 & 44.4 & 83.0 & \textbf{83.0} & 0.830 \\
PsyCIRA & PsyDARA & 78.9 & 100.0 & 100.0 & 55.6 & \textbf{84.9} & \textbf{83.0} & \textbf{0.840} \\
\addlinespace[1.5pt]
\multicolumn{9}{@{}l}{\textit{GPT-5.6 Luna}} \\
Direct & Direct & 78.9 & 73.3 & 70.0 & 22.2 & 66.0 & 62.3 & 0.642 \\
Direct & PsyDARA & 73.7 & 100.0 & 70.0 & 44.4 & 75.5 & 71.7 & 0.736 \\
PsyCIRA & Direct & 89.5 & 73.3 & 70.0 & 22.2 & 69.8 & 64.2 & 0.667 \\
PsyCIRA & PsyDARA & 89.5 & 100.0 & 80.0 & 55.6 & \textbf{84.9} & 77.4 & 0.811 \\
\addlinespace[1.5pt]
\multicolumn{9}{@{}l}{\textit{DeepSeek V4 Flash}} \\
Direct & Direct & 63.2 & 93.3 & 80.0 & 33.3 & 69.8 & 64.2 & 0.667 \\
Direct & PsyDARA & 63.2 & 100.0 & 80.0 & 22.2 & 69.8 & 67.9 & 0.689 \\
PsyCIRA & Direct & 68.4 & 93.3 & 100.0 & 33.3 & 75.5 & 71.7 & 0.736 \\
PsyCIRA & PsyDARA & 73.7 & 100.0 & 90.0 & 44.4 & 79.2 & 71.7 & 0.755 \\
\addlinespace[1.5pt]
\multicolumn{9}{@{}l}{\textit{DeepSeek V4 Pro}} \\
Direct & Direct & 63.2 & 93.3 & 70.0 & 33.3 & 67.9 & 64.2 & 0.660 \\
Direct & PsyDARA & 63.2 & 93.3 & 60.0 & 33.3 & 66.0 & 64.2 & 0.651 \\
PsyCIRA & Direct & 68.4 & 100.0 & 90.0 & 44.4 & 77.4 & 75.5 & 0.764 \\
PsyCIRA & PsyDARA & 63.2 & 100.0 & 90.0 & 55.6 & 77.4 & 75.5 & 0.764 \\
\bottomrule
\end{tabular*}%
\endgroup

  \caption{Simulated patients: evaluation cohort (53 cases, 16 exchanges). OC-related: obsessive-compulsive or related disorders; Other: seven pooled categories. Bold: table-wide maxima for overall metrics, including ties.}
  \label{tab:evaluation}
\end{table*}

\subsection{Evaluation measures}
Diagnostic agreement uses exact ICD-11 base-code matches to the source-case reference in simulation and the reconciled expert primary code in humans. \emph{Rank-1 accuracy} (R1) requires the reference first among the zero to three final hypotheses; \emph{coverage} accepts it anywhere; \emph{mean reciprocal rank} (MRR) averages its reciprocal position, assigning zero when absent. Mean list length averages the number of final hypotheses. Human \emph{diagnostic decision agreement} measures agreement with experts on whether to propose a diagnosis. Code metrics use transcripts with an expert-proposed primary diagnosis: 23 in the Direct arm, 20 in the PsyCIDRA arm, or all 43 for shared-transcript comparison. A decision not to propose a diagnosis can reflect a presentation that does not meet diagnostic criteria or insufficient information for a diagnostic conclusion.

We use a 12-item text adaptation (0--3) of the validated \emph{Clinical Communication Skills Scale--Short Version} (CCSS-S) \citep{maass-2022-ccss-s}. For AI interviewing, psychologists developed the \emph{AI Interview Examination} (AIIE; 14 items, 1--5) to assess diagnostic inquiry, risk recognition, rapport, and interview flow, informed by clinical-dialogue evaluation \citep{tu-2025-amie,johri-2025-craft-md}. They also developed the \emph{AI Safety Check} (AISC; seven binary items) for crisis handling, role boundaries, advice, relational behavior, and data requests, informed by health-AI guidance and mental-health risks \citep{who-2024-lmm-guidance,lawrence-2024-llm-mental-health-risks}.

Quality scores average items, then available reviewers within transcripts and transcripts within conditions; unassessable CCSS-S items are omitted. Any reviewer flag counts as an AISC concern, separately from live safety stops.

\section{Results}
\label{sec:results}

\subsection{Simulated patients}
On the evaluation cohort, PsyCIDRA improves R1 over Direct by 7.5--15.1 percentage points and coverage by 5.7--18.9 points across all four model settings (Table~\ref{tab:evaluation}). These gains build on already substantial agreement under Direct prompting.

With Terra and both DeepSeek variants, most of the R1 gain follows the change to PsyCIRA interviews; Luna benefits more strongly from PsyDARA under either interviewer. The lower-cost Luna setting has a larger overall gain, halving its R1 gap with Terra and matching Terra's coverage \citep{openai-gpt56-luna,openai-gpt56-terra}.

Higher agreement also comes with more selective outputs: with Terra, PsyCIDRA raises coverage while reducing mean list length from 1.26 to 1.08. Yet these gains are uneven across presentations. PsyCIDRA covers every OC-related reference across all four backbones, whereas coverage in the heterogeneous Other group remains around half. The contrast points to diagnostic breadth as an unresolved challenge even when aggregate agreement is high.

PsyCIRA also receives higher expert ratings than Direct on overall AIIE (3.98 versus 3.73/5) and adapted CCSS-S (2.62 versus 2.39/3). The AIIE differences concentrate in rapport and flow, while diagnostic-inquiry ratings remain similar. Thus, the higher overall ratings primarily reflect how the conversation is conducted. AISC concerns occur in one of 23 PsyCIRA transcripts and none of 23 Direct transcripts, keeping safety a separate evaluation target.

\begin{figure}[t]
  \centering
  \includegraphics[width=\columnwidth]{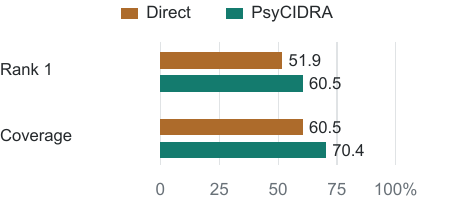}
  \caption{Simulated patients: diagnostic agreement on 81 held-out cases.}
  \label{fig:heldout}
\end{figure}

PsyCIDRA retains its advantage on held-out cases, reaching 60.5\% R1 versus 51.9\% for Direct and 70.4\% coverage versus 60.5\% (Figure~\ref{fig:heldout}). Both systems fall below their evaluation-cohort results. This decrease may partly reflect the test's broader range of presentations: 44/81 cases fall outside the anxiety/fear, OC-related, and mood groups, compared with 9/53 during evaluation.

\subsection{Human participants}
\begin{figure}[t]
  \centering
  \includegraphics[width=\columnwidth]{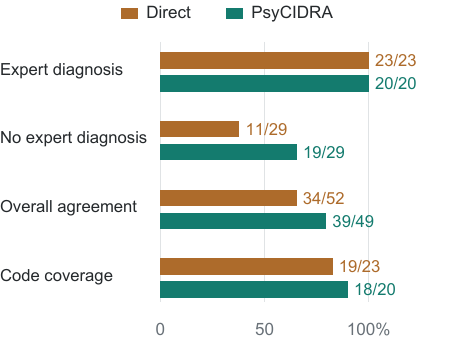}
  \caption{Human participants: diagnostic decision agreement and code coverage. Direct: $n=52$; PsyCIDRA: $n=49$. Coverage uses transcripts with an expert-proposed diagnosis (23 and 20, respectively).}
  \label{fig:human-alignment}
\end{figure}

In the human study, PsyCIDRA agrees more often with experts on whether the transcript supports a final hypothesis: 79.6\% versus 65.4\% for Direct (Figure~\ref{fig:human-alignment}). Among transcripts with an expert-proposed diagnosis, reference-code coverage is also higher (90.0\% versus 82.6\%); the larger separation concerns withholding a hypothesis when experts also do so. Human assessment thus adds a decision about diagnostic sufficiency to code identification. These are transcript-based judgments in separate participant arms.

Comparing diagnosers on shared transcripts makes this pattern clearer. Among the 43 transcripts with an expert-proposed diagnosis, PsyDARA ranks the primary code first in 29 cases versus 26 for Direct, while coverage differs by only one case (37 versus 36). Across the 58 transcripts where experts propose no diagnosis, PsyDARA also withholds a hypothesis in 39 cases versus 27 for Direct. The main change is therefore in deciding whether to issue a final hypothesis, with a smaller change in identifying the reference code. PsyDARA also withholds a hypothesis on one transcript with an expert-proposed diagnosis, illustrating why this decision must be assessed alongside diagnostic coverage.

Interview ratings again distinguish conversational ease from clinical coverage. Overall AIIE scores are close (3.75/5 for Direct, 3.83/5 for PsyCIRA), while adapted CCSS-S favors PsyCIRA (2.65/3 versus 2.51/3 for Direct). PsyCIRA receives higher rapport and flow ratings, but lower diagnostic-inquiry ratings (3.53 versus 3.68/5). This pattern identifies more thorough clinical inquiry as a priority alongside the ability to sustain an organized, supportive exchange.

Safety remains an active concern within these otherwise promising interactions. AISC flags occur in 2/52 Direct and 3/49 PsyCIRA transcripts, including crisis-handling concerns in two PsyCIRA interviews. Separately, one analyzed session in each arm activates the live safety stop; four further stopped sessions were excluded (Appendix~\ref{app:human-attrition}). The checklist findings identify behavior requiring improvement, while the stops document use of the intervention mechanism; both matter when assessing readiness for further clinical study. These counts describe the analyzed cohort and do not characterize safety across all registrants.

\section{Conclusion}
\label{sec:discussion}

This feasibility study provides initial evidence that LLMs can conduct free-form psychiatric interviews and derive hypotheses aligned with expert transcript review. Direct prompting already gives promising results, while PsyCIDRA has higher observed diagnostic agreement and some interview-quality ratings. In humans, the clearest diagnostic-stage difference concerns whether the available evidence warrants a final hypothesis. These findings coexist with diagnostic errors, incomplete inquiry, and safety concerns. Such systems should serve only as assistants to qualified experts, who must review their evidence and hypotheses and retain responsibility for diagnostic decisions. Dedicated safety evaluation and crisis safeguards remain necessary. Prospective clinical studies with independent diagnostic assessments should test whether these tools improve experts' decisions and how failures are detected and managed across presentations and care settings.

\label{lastcontentpage}
\section*{Limitations}
\label{limitationspage}

The human reference was established by two psychologists who independently reviewed the interview transcripts and reconciled their primary judgments. However, both experts and diagnostic agents relied on information elicited by the interviewer. The experts did not conduct independent clinical interviews, obtain collateral histories, or assess participants longitudinally. Agreement therefore measures consistency with expert interpretation of the available transcript; shared omissions may produce agreement without establishing diagnostic correctness.

Text-only interaction also excludes direct observation of appearance, facial expression, psychomotor behavior, and vocal characteristics. Participants may describe these features, but self-report does not replace clinical observation. Similar limitations of text-based assessment have been identified in medical and mental-health conversational-agent research \citep{tu-2025-amie,sikstrom-2025-talktoalba}.

The human study was a feasibility evaluation involving self-selected adult volunteers in Iran. Demographic diversity does not make this cohort representative of the general population or a clinically stratified sample. Enrollment and interview completion introduce selection effects. The complete-system comparisons involve different participants, limiting attribution of observed differences to the systems. The cohort cannot establish population performance or the frequency of rare harms.

Simulation provides controlled comparisons but does not replace evaluation with real patients. Simulated presentations may not capture the complexity and variability of clinical encounters.

Finally, structured reports do not establish faithful explanations or clinical benefit. We did not evaluate improvements in clinicians' decisions, review time, or patient outcomes. Prospective evaluation with independent clinical references and broader recruitment is necessary before drawing conclusions about clinical effectiveness.

\section*{Ethical considerations}
\label{ethicspage}

\ifdefined\PsyCIDRAReview
Ethics-board approval was obtained for the human study. The institution and official approval identifier are withheld for double-blind review and can be provided in the nonanonymous version.
\else
The human study was approved by the Research Ethics Committee of the Faculty of Psychology and Educational Sciences, University of Tehran (approval code \href{https://ethics.research.ac.ir/IR.UT.PSYEDU.REC.1404.138}{IR.UT.PSYEDU.REC.1404.138}).
\fi
Adult volunteers gave online informed consent and participated voluntarily without compensation. They could skip questions or discontinue participation, and were invited to complete post-interview mental-health self-report and survey forms. Individual diagnostic reports were not returned during the evaluated study.

Registration did not request direct identifying information, but mental-health narratives can contain sensitive or indirectly identifying details. Human-participant dialogues will not be publicly released. Release of simulation materials is subject to source-data permissions and privacy review, including details retained in deidentified records.

Participants had a control to terminate the chat if they experienced discomfort or needed urgent help. The interviewers were instructed to stop ordinary interviewing when acute danger was identified, display a fixed message directing participants toward human or emergency support, and record the event for human review. The retained records establish referral messages and pending-review flags, but not response times or completed clinical follow-up. These safeguards did not guarantee crisis detection or immediate intervention. Automatic diagnosis was suppressed after a live safety stop; the two stopped transcripts included in this analysis were processed separately offline for research.

The study evaluates technical performance in psychiatric interviewing and diagnostic reasoning. It does not establish readiness for clinical deployment, including under supervision. Neither interviewer was intended to provide diagnoses, treatment recommendations, or prescribing advice to participants. A proposed hypothesis is not a confirmed diagnosis, and withholding a hypothesis does not establish absence of a disorder. Empathetic language may encourage trust beyond the evidence \citep{lawrence-2024-llm-mental-health-risks}.

Few observed safety flags do not establish clinical safety. Even isolated failures, particularly missed crisis responses, warrant investigation. Deployment requires dedicated safety evaluation, independent clinical validation, assessment across demographic and linguistic groups, and appropriate clinical governance.

\bibliography{references,references-appendix}

\begin{thebibliography}{55}
\providecommand{\natexlab}[1]{#1}

\bibitem[{Abd-alrazaq et~al.(2022)Abd-alrazaq, Alhuwail, Schneider, Toro,
  Ahmed, Alzubaidi, Alajlani, and Househ}]{abd-alrazaq-2022-ai-diagnosis}
Alaa Abd-alrazaq, Dari Alhuwail, Jens Schneider, Carla~T. Toro, Arfan Ahmed,
  Mahmood Alzubaidi, Mohannad Alajlani, and Mowafa Househ. 2022.
\newblock \href {https://doi.org/10.1038/s41746-022-00631-8} {The performance
  of artificial intelligence-driven technologies in diagnosing mental
  disorders: An umbrella review}.
\newblock \emph{npj Digital Medicine}, 5:87.

\bibitem[{Adeponle et~al.(2012)Adeponle, Thombs, Groleau, Jarvis, and
  Kirmayer}]{adeponle-2012-cultural-formulation}
Ademola~B. Adeponle, Brett~D. Thombs, Danielle Groleau, Eric Jarvis, and
  Laurence~J. Kirmayer. 2012.
\newblock \href {https://doi.org/10.1176/appi.ps.201100280} {Using the cultural
  formulation to resolve uncertainty in diagnoses of psychosis among
  ethnoculturally diverse patients}.
\newblock \emph{Psychiatric Services}, 63(2):147--153.

\bibitem[{AlKhamissi et~al.(2024)AlKhamissi, ElNokrashy, Alkhamissi, and
  Diab}]{alkhamissi-2024-cultural-alignment}
Badr AlKhamissi, Muhammad ElNokrashy, Mai Alkhamissi, and Mona Diab. 2024.
\newblock \href {https://doi.org/10.18653/v1/2024.acl-long.671} {Investigating
  cultural alignment of large language models}.
\newblock In \emph{Proceedings of the 62nd Annual Meeting of the Association
  for Computational Linguistics (Volume 1: Long Papers)}, pages 12404--12422,
  Bangkok, Thailand. Association for Computational Linguistics.

\bibitem[{Bi et~al.(2025)Bi, Chen, Liu, Wang, Xiao, Xie, Zhang, Huang, Chen,
  Peng, and Huang}]{bi-2025-magi}
Guanqun Bi, Zhuang Chen, Zhoufu Liu, Hongkai Wang, Xiyao Xiao, Yuqiang Xie, Wen
  Zhang, Yongkang Huang, Yuxuan Chen, Libiao Peng, and Minlie Huang. 2025.
\newblock \href {https://doi.org/10.18653/v1/2025.findings-acl.1278} {{MAGI}:
  Multi-agent guided interview for psychiatric assessment}.
\newblock In \emph{Findings of the Association for Computational Linguistics:
  ACL 2025}, pages 24898--24921, Vienna, Austria. Association for Computational
  Linguistics.

\bibitem[{Bi et~al.(2026)Bi, Liu, Chen, Wan, Xiao, and Huang}]{bi-2026-s4}
Guanqun Bi, Zhoufu Liu, Zhuang Chen, Dazhen Wan, Xiyao Xiao, and Minlie Huang.
  2026.
\newblock \href {https://doi.org/10.18653/v1/2026.acl-long.1802} {{S$^4$}:
  Operationalizing speech act theory for strategic semi-structured psychiatric
  interview}.
\newblock In \emph{Proceedings of the 64th Annual Meeting of the {A}ssociation
  for {C}omputational {L}inguistics (Volume 1: Long Papers)}, pages
  38881--38899, San Diego, California, United States. Association for
  Computational Linguistics.

\bibitem[{Cao et~al.(2025)Cao, Huang, Zhang, Deng, Zhang, and
  Lei}]{cao-2025-stigma}
Jieming Cao, Chen Huang, Yanan Zhang, Ruibo Deng, Jincheng Zhang, and Wenqiang
  Lei. 2025.
\newblock \href {https://doi.org/10.18653/v1/2025.findings-naacl.10} {Breaking
  the stigma! unobtrusively probe symptoms in depression disorder diagnosis
  dialogue}.
\newblock In \emph{Findings of the Association for Computational Linguistics:
  NAACL 2025}, pages 182--200, Albuquerque, New Mexico. Association for
  Computational Linguistics.

\bibitem[{Cook et~al.(2025)Cook, Overgaard, Pankratz, Del~Fiol, and
  Aakre}]{cook-2025-virtual-patients}
David~A. Cook, Joshua Overgaard, V.~Shane Pankratz, Guilherme Del~Fiol, and
  Chris~A. Aakre. 2025.
\newblock \href {https://doi.org/10.2196/68486} {Virtual patients using large
  language models: Scalable, contextualized simulation of clinician-patient
  dialogue with feedback}.
\newblock \emph{Journal of Medical Internet Research}, 27:e68486.

\bibitem[{{DeepSeek}(2026)}]{deepseek-v4-changelog}
{DeepSeek}. 2026.
\newblock \href {https://api-docs.deepseek.com/updates/} {Change log:
  {DeepSeek-V4} {API} releases}.
\newblock API documentation. Accessed October 3, 2026.

\bibitem[{Dwyer et~al.(2018)Dwyer, Falkai, and
  Koutsouleris}]{dwyer-2018-machine-learning}
Dominic~B. Dwyer, Peter Falkai, and Nikolaos Koutsouleris. 2018.
\newblock \href {https://doi.org/10.1146/annurev-clinpsy-032816-045037}
  {Machine learning approaches for clinical psychology and psychiatry}.
\newblock \emph{Annual Review of Clinical Psychology}, 14:91--118.

\bibitem[{First et~al.(2016)First, Williams, Karg, and
  Spitzer}]{first-2016-scid5-cv}
Michael~B. First, Janet B.~W. Williams, Rhonda~S. Karg, and Robert~L. Spitzer.
  2016.
\newblock \emph{Structured Clinical Interview for {DSM-5} Disorders, Clinician
  Version ({SCID-5-CV})}.
\newblock American Psychiatric Association Publishing, Arlington, VA.

\bibitem[{Garc{\'i}a-Torres et~al.(2025)Garc{\'i}a-Torres, Fern{\'a}ndez, Mira,
  Morales, and Vicente}]{garcia-torres-2025-vsp}
Daniel Garc{\'i}a-Torres, C{\'e}sar Fern{\'a}ndez, Jos{\'e}~Joaqu{\'i}n Mira,
  Alexandra Morales, and Mar{\'i}a~Asunci{\'o}n Vicente. 2025.
\newblock \href {https://doi.org/10.2196/78857} {Using {AI}-based virtual
  simulated patients for training in psychopathological interviewing:
  Cross-sectional observational study}.
\newblock \emph{JMIR Medical Education}, 11:e78857.

\bibitem[{Holmes et~al.(2011)Holmes, Singh, and
  McColl}]{holmes-2011-hypothesis-driven}
Alex Holmes, Bruce Singh, and Geoff McColl. 2011.
\newblock \href {https://doi.org/10.3109/10398562.2011.601462} {Revisiting the
  hypothesis-driven interview in a contemporary context}.
\newblock \emph{Australasian Psychiatry}, 19(6):484--488.

\bibitem[{Hu et~al.(2025)Hu, Chen, Chen, Mu, Shao, and Luo}]{hu-2025-hiagent}
Mengkang Hu, Tianxing Chen, Qiguang Chen, Yao Mu, Wenqi Shao, and Ping Luo.
  2025.
\newblock \href {https://doi.org/10.18653/v1/2025.acl-long.1575} {{HiAgent}:
  Hierarchical working memory management for solving long-horizon agent tasks
  with large language model}.
\newblock In \emph{Proceedings of the 63rd Annual Meeting of the Association
  for Computational Linguistics (Volume 1: Long Papers)}, pages 32779--32798,
  Vienna, Austria. Association for Computational Linguistics.

\bibitem[{Iman and Conover(1982)}]{iman-1982-rank-correlation}
Ronald~L. Iman and W.~J. Conover. 1982.
\newblock \href {https://doi.org/10.1080/03610918208812265} {A
  distribution-free approach to inducing rank correlation among input
  variables}.
\newblock \emph{Communications in Statistics---Simulation and Computation},
  11(3):311--334.

\bibitem[{Johri et~al.(2025)Johri, Jeong, Tran, Schlessinger, Wongvibulsin,
  Barnes, Zhou, Cai, Van~Allen, Kim, Daneshjou, and
  Rajpurkar}]{johri-2025-craft-md}
Shreya Johri, Jaehwan Jeong, Benjamin~A. Tran, Daniel~I. Schlessinger, Shannon
  Wongvibulsin, Leandra~A. Barnes, Hong-Yu Zhou, Zhuo~Ran Cai, Eliezer~M.
  Van~Allen, David Kim, Roxana Daneshjou, and Pranav Rajpurkar. 2025.
\newblock \href {https://doi.org/10.1038/s41591-024-03328-5} {An evaluation
  framework for clinical use of large language models in patient interaction
  tasks}.
\newblock \emph{Nature Medicine}, 31(1):77--86.

\bibitem[{Kyung et~al.(2025)Kyung, Chung, Bae, Kim, Sohn, Kim, Kim, and
  Choi}]{kyung-2025-patientsim}
Daeun Kyung, Hyunseung Chung, Seongsu Bae, Jiho Kim, Jae~Ho Sohn, Taerim Kim,
  Soo~Kyung Kim, and Edward Choi. 2025.
\newblock \href {https://doi.org/10.52202/085713-0861} {{PatientSim}: A
  persona-driven simulator for realistic doctor-patient interactions}.
\newblock In \emph{Advances in Neural Information Processing Systems},
  volume~38.

\bibitem[{Lawrence et~al.(2024)Lawrence, Schneider, Rubin, Matari{\'c}, McDuff,
  and Jones~Bell}]{lawrence-2024-llm-mental-health-risks}
Hannah~R. Lawrence, Renee~A. Schneider, Susan~B. Rubin, Maja~J. Matari{\'c},
  Daniel~J. McDuff, and Megan Jones~Bell. 2024.
\newblock \href {https://doi.org/10.2196/59479} {The opportunities and risks of
  large language models in mental health}.
\newblock \emph{JMIR Mental Health}, 11:e59479.

\bibitem[{Lee et~al.(2025)Lee, Lim, Jung, and Kim}]{lee-2025-psyche}
Jingoo Lee, Kyungho Lim, Young-Chul Jung, and Byung-Hoon Kim. 2025.
\newblock \href {https://doi.org/10.48550/arXiv.2501.01594} {{PSYCHE}: A
  multi-faceted patient simulation framework for evaluation of psychiatric
  assessment conversational agents}.
\newblock \emph{Preprint}, arXiv:2501.01594.

\bibitem[{Lewis-Fern{\'a}ndez et~al.(2014)Lewis-Fern{\'a}ndez, Aggarwal,
  B{\"a}{\"a}rnhielm, Rohlof, Kirmayer, Weiss, Jadhav, Hinton, Alarc{\'o}n,
  Bhugra, Groen, van Dijk, Qureshi, Collazos, Rousseau, Caballero, Ramos, and
  Lu}]{lewis-fernandez-2014-cultural-formulation}
Roberto Lewis-Fern{\'a}ndez, Neil~Krishan Aggarwal, Sofie B{\"a}{\"a}rnhielm,
  Hans Rohlof, Laurence~J. Kirmayer, Mitchell~G. Weiss, Sushrut Jadhav, Ladson
  Hinton, Renato~D. Alarc{\'o}n, Dinesh Bhugra, Simon Groen, Rob van Dijk, Adil
  Qureshi, Francisco Collazos, C{\'e}cile Rousseau, Luis Caballero, Mar Ramos,
  and Francis Lu. 2014.
\newblock \href {https://doi.org/10.1521/psyc.2014.77.2.130} {Culture and
  psychiatric evaluation: Operationalizing cultural formulation for {DSM-5}}.
\newblock \emph{Psychiatry}, 77(2):130--154.

\bibitem[{Li{\'e}vin et~al.(2026)Li{\'e}vin, Schmidgall, Strother, Bijamov,
  Goel, Palepu, Park, Balazadeh, Sun, Guerard, Chen, Steiner, Dhillon, Azar,
  Mehta, Spetsieris, Shah, Abdelrahim, Dahiya, Liu, Chou, Matias, Hassidim,
  Webster, Le, Hadsell, Barral, Radebaugh, Faust, Azizi, Schaekermann, Chen,
  Tu, Racz, and Yang}]{lievin-2026-residencyrl}
Valentin Li{\'e}vin, Samuel Schmidgall, Tim Strother, Alex Bijamov, Akshay
  Goel, Anil Palepu, Chunjong Park, Vahid Balazadeh, Min~Woo Sun, Marius
  Guerard, Justin Chen, Dave Steiner, Vikram Dhillon, Ibrahim Azar, Akhil
  Mehta, Nicholas Spetsieris, Shilpan Shah, Maen Abdelrahim, Amit Dahiya, and
  16 others. 2026.
\newblock \href {https://doi.org/10.48550/arXiv.2608.07418} {{ResidencyRL}:
  Reinforcement learning in simulated clinical environments}.
\newblock \emph{Preprint}, arXiv:2608.07418.

\bibitem[{Louie et~al.(2024)Louie, Nandi, Fang, Chang, Brunskill, and
  Yang}]{louie-2024-roleplay-doh}
Ryan Louie, Ananjan Nandi, William Fang, Cheng Chang, Emma Brunskill, and Diyi
  Yang. 2024.
\newblock \href {https://doi.org/10.18653/v1/2024.emnlp-main.591}
  {{Roleplay-doh}: Enabling domain-experts to create {LLM}-simulated patients
  via eliciting and adhering to principles}.
\newblock In \emph{Proceedings of the 2024 Conference on Empirical Methods in
  Natural Language Processing}, pages 10570--10603, Miami, Florida, USA.
  Association for Computational Linguistics.

\bibitem[{Maa{\ss} et~al.(2022)Maa{\ss}, K{\"u}hne, Heinze, Ay-Bryson, and
  Weck}]{maass-2022-ccss-s}
Ulrike Maa{\ss}, Franziska K{\"u}hne, Peter~Eric Heinze, Destina~Sevde
  Ay-Bryson, and Florian Weck. 2022.
\newblock \href {https://doi.org/10.3389/fpsyt.2022.977324} {The concise
  measurement of clinical communication skills: Validation of a short scale}.
\newblock \emph{Frontiers in Psychiatry}, 13:977324.

\bibitem[{Merrill and Sabharwal(2024)}]{merrill-2024-expressive}
William Merrill and Ashish Sabharwal. 2024.
\newblock \href
  {https://proceedings.iclr.cc/paper_files/paper/2024/hash/1f59721c106ea80f613299039112f651-Abstract-Conference.html}
  {The expressive power of transformers with chain of thought}.
\newblock In \emph{The Twelfth International Conference on Learning
  Representations}.

\bibitem[{Mi et~al.(2026)Mi, Ma, Yang, Li, Wang, Zhang, and
  Wang}]{mi-2026-skill-pro}
Qirui Mi, Zhijian Ma, Mengyue Yang, Haoxuan Li, Yisen Wang, Haifeng Zhang, and
  Jun Wang. 2026.
\newblock \href {https://proceedings.mlr.press/v306/mi26d.html} {{Skill-Pro}:
  Learning reusable skills from experience via non-parametric {PPO} for {LLM}
  agents}.
\newblock In \emph{Proceedings of the 43rd International Conference on Machine
  Learning}, volume 306 of \emph{Proceedings of Machine Learning Research},
  pages 88574--88595. PMLR.

\bibitem[{Naous et~al.(2024)Naous, Ryan, Ritter, and
  Xu}]{naous-2024-cultural-bias}
Tarek Naous, Michael~J. Ryan, Alan Ritter, and Wei Xu. 2024.
\newblock \href {https://doi.org/10.18653/v1/2024.acl-long.862} {Having beer
  after prayer? measuring cultural bias in large language models}.
\newblock In \emph{Proceedings of the 62nd Annual Meeting of the Association
  for Computational Linguistics (Volume 1: Long Papers)}, pages 16366--16393,
  Bangkok, Thailand. Association for Computational Linguistics.

\bibitem[{{OpenAI}(2026{\natexlab{a}})}]{openai-gpt56-luna}
{OpenAI}. 2026{\natexlab{a}}.
\newblock \href {https://developers.openai.com/api/docs/models/gpt-5.6-luna}
  {{GPT-5.6 Luna} model}.
\newblock API documentation. Accessed October 3, 2026.

\bibitem[{{OpenAI}(2026{\natexlab{b}})}]{openai-gpt56-terra}
{OpenAI}. 2026{\natexlab{b}}.
\newblock \href {https://developers.openai.com/api/docs/models/gpt-5.6-terra}
  {{GPT-5.6 Terra} model}.
\newblock API documentation. Accessed October 3, 2026.

\bibitem[{Park et~al.(2023)Park, O'Brien, Cai, Morris, Liang, and
  Bernstein}]{park-2023-generative-agents}
Joon~Sung Park, Joseph~C. O'Brien, Carrie~J. Cai, Meredith~Ringel Morris, Percy
  Liang, and Michael~S. Bernstein. 2023.
\newblock \href {https://doi.org/10.1145/3586183.3606763} {Generative agents:
  Interactive simulacra of human behavior}.
\newblock In \emph{Proceedings of the 36th Annual ACM Symposium on User
  Interface Software and Technology}, pages 1--22. Association for Computing
  Machinery.

\bibitem[{Rollwage et~al.(2026)Rollwage, McFadyen, Juchems, Balogh, Pisupati,
  Mircea, Hauser, Prichard, and Harper}]{rollwage-2026-cognitive-layer}
Max Rollwage, Jessica McFadyen, Keno Juchems, Annamaria Balogh, Sashank
  Pisupati, Margareta-Theodora Mircea, Tobias~U. Hauser, George Prichard, and
  Ross Harper. 2026.
\newblock \href {https://doi.org/10.1038/s41591-026-04278-w} {A cognitive layer
  architecture to support large-language model performance in psychotherapy
  interactions}.
\newblock \emph{Nature Medicine}, 32(5):1717--1725.

\bibitem[{Rosenman et~al.(2024)Rosenman, Wolf, and
  Hendler}]{rosenman-2024-questionnaire}
Gony Rosenman, Lior Wolf, and Talma Hendler. 2024.
\newblock \href {https://doi.org/10.18653/v1/2024.findings-emnlp.23} {{LLM}
  questionnaire completion for automatic psychiatric assessment}.
\newblock In \emph{Findings of the Association for Computational Linguistics:
  EMNLP 2024}, pages 403--415, Miami, Florida, USA. Association for
  Computational Linguistics.

\bibitem[{Santomauro et~al.(2024)Santomauro, Vos, Whiteford, Chisholm, Saxena,
  and Ferrari}]{santomauro-2024-treatment}
Damian~F. Santomauro, Theo Vos, Harvey~A. Whiteford, Dan Chisholm, Shekhar
  Saxena, and Alize~J. Ferrari. 2024.
\newblock \href {https://doi.org/10.1016/S2215-0366(24)00317-1} {Service
  coverage for major depressive disorder: Estimated rates of minimally adequate
  treatment for 204 countries and territories in 2021}.
\newblock \emph{The Lancet Psychiatry}, 11(12):1012--1021.

\bibitem[{Savander et~al.(2021)Savander, Hintikka, Wuolio, and
  Per{\"a}kyl{\"a}}]{savander-2021-subjective-experiences}
Enik{\"o}~{\`E}va Savander, Jukka Hintikka, Mariel Wuolio, and Anssi
  Per{\"a}kyl{\"a}. 2021.
\newblock \href {https://doi.org/10.3389/fpsyt.2021.605760} {The patients'
  practises disclosing subjective experiences in the psychiatric intake
  interview}.
\newblock \emph{Frontiers in Psychiatry}, 12:605760.

\bibitem[{Schlager et~al.(2026)Schlager, Jungmann, Schmidgall, Raffler, Hartl,
  Wende, Ro{\ss}m{\"u}ller, Ketzer, Hassidim, Webster, Matias, Liu, Rueckert,
  Schaekermann, and Hager}]{schlager-2026-pwp}
Moritz Schlager, Friederike Jungmann, Samuel Schmidgall, Philipp Raffler,
  Franziska Hartl, Eva Wende, Paula Ro{\ss}m{\"u}ller, Conrad Ketzer, Avinatan
  Hassidim, Dale~R. Webster, Yossi Matias, Yun Liu, Daniel Rueckert, Mike
  Schaekermann, and Paul Hager. 2026.
\newblock \href {https://doi.org/10.48550/arXiv.2606.17441} {Patients with
  personality: Realistic patient simulation through controlled diversity and
  selective disclosure}.
\newblock \emph{Preprint}, arXiv:2606.17441.
\newblock Version 2.

\bibitem[{Shatte et~al.(2019)Shatte, Hutchinson, and
  Teague}]{shatte-2019-machine-learning}
Adrian B.~R. Shatte, Delyse~M. Hutchinson, and Samantha~J. Teague. 2019.
\newblock \href {https://doi.org/10.1017/S0033291719000151} {Machine learning
  in mental health: A scoping review of methods and applications}.
\newblock \emph{Psychological Medicine}, 49(9):1426--1448.

\bibitem[{Sheehan et~al.(1998)Sheehan, Lecrubier, Weiller, Sheehan, Amorim,
  Janavs, Hergueta, Baker, and Dunbar}]{sheehan-1998-mini}
David~V. Sheehan, Yves Lecrubier, Emmanuelle Weiller, K.~Harnett Sheehan,
  Patricia Amorim, Juris Janavs, Thierry Hergueta, Roxy Baker, and Geoffrey~C.
  Dunbar. 1998.
\newblock \href
  {https://www.psychiatrist.com/jcp/mini-international-neuropsychiatric-interview-mini/}
  {The {Mini-International Neuropsychiatric Interview (M.I.N.I.)}: The
  development and validation of a structured diagnostic psychiatric interview
  for {DSM-IV} and {ICD-10}}.
\newblock \emph{The Journal of Clinical Psychiatry}, 59(Suppl. 20):22--33.

\bibitem[{Shinn et~al.(2023)Shinn, Cassano, Gopinath, Narasimhan, and
  Yao}]{shinn-2023-reflexion}
Noah Shinn, Federico Cassano, Ashwin Gopinath, Karthik Narasimhan, and Shunyu
  Yao. 2023.
\newblock \href {https://doi.org/10.52202/075280-0377} {{Reflexion}: Language
  agents with verbal reinforcement learning}.
\newblock In \emph{Advances in Neural Information Processing Systems},
  volume~36, pages 8634--8652. Curran Associates, Inc.

\bibitem[{Sikstr{\"o}m et~al.(2025)Sikstr{\"o}m, Boehme, Mirstr{\"o}m,
  Agbotsoka, Gy{\H{o}}ri, Lasota, Tabesh, Stille, and
  Garcia}]{sikstrom-2025-talktoalba}
Sverker Sikstr{\"o}m, Rebecca~Astrid Boehme, Mariam Mirstr{\"o}m, Thibaud
  Agbotsoka, Gerg{\H{o}} Gy{\H{o}}ri, Marta Lasota, Mona Tabesh, Lotta Stille,
  and Danilo Garcia. 2025.
\newblock \href {https://doi.org/10.1038/s41598-025-13429-x} {Generative
  {AI}-assisted clinical interviewing of mental health}.
\newblock \emph{Scientific Reports}, 15:37737.

\bibitem[{Singhal et~al.(2025)Singhal, Tu, Gottweis, Sayres, Wulczyn, Amin,
  Hou, Clark, Pfohl, Cole-Lewis, Neal, Rashid, Schaekermann, Wang, Dash, Chen,
  Shah, Lachgar, Mansfield, Prakash, Green, Dominowska, Ag{\"u}era~y Arcas,
  Toma{\v{s}}ev, Liu, Wong, Semturs, Mahdavi, Barral, Webster, Corrado, Matias,
  Azizi, Karthikesalingam, and Natarajan}]{singhal-2025-medpalm2}
Karan Singhal, Tao Tu, Juraj Gottweis, Rory Sayres, Ellery Wulczyn, Mohamed
  Amin, Le~Hou, Kevin Clark, Stephen~R. Pfohl, Heather Cole-Lewis, Darlene
  Neal, Qazi~Mamunur Rashid, Mike Schaekermann, Amy Wang, Dev Dash, Jonathan~H.
  Chen, Nigam~H. Shah, Sami Lachgar, Philip~Andrew Mansfield, and 16 others.
  2025.
\newblock \href {https://doi.org/10.1038/s41591-024-03423-7} {Toward
  expert-level medical question answering with large language models}.
\newblock \emph{Nature Medicine}, 31:943--950.

\bibitem[{Soto and John(2017)}]{soto-2017-bfi2}
Christopher~J. Soto and Oliver~P. John. 2017.
\newblock \href {https://doi.org/10.1037/pspp0000096} {The next {Big Five
  Inventory} ({BFI-2}): Developing and assessing a hierarchical model with 15
  facets to enhance bandwidth, fidelity, and predictive power}.
\newblock \emph{Journal of Personality and Social Psychology}, 113(1):117--143.

\bibitem[{Sun et~al.(2026)Sun, Yang, Jiang, Tian, Zhu, Zhong, Lei, Huang, Zeng,
  Zhou, Xiao, and Wei}]{sun-2026-mentalseek-dx}
Xiao Sun, Yuming Yang, Xinyi Jiang, Yu~Tian, Junnan Zhu, Jiang Zhong, Qin Lei,
  Jingwang Huang, Haoyang Zeng, Xinyu Zhou, Xin Xiao, and Kaiwen Wei. 2026.
\newblock \href {https://doi.org/10.18653/v1/2026.acl-long.1225}
  {{MentalSeek-Dx}: Towards progressive hypothetico-deductive reasoning for
  real-world psychiatric diagnosis}.
\newblock In \emph{Proceedings of the 64th Annual Meeting of the Association
  for Computational Linguistics (Volume 1: Long Papers)}, pages 26600--26636,
  San Diego, California, United States. Association for Computational
  Linguistics.

\bibitem[{Tu et~al.(2025)Tu, Schaekermann, Palepu, Saab, Freyberg, Tanno, Wang,
  Li, Amin, Cheng, Vedadi, Tomasev, Azizi, Singhal, Hou, Webson, Kulkarni,
  Mahdavi, Semturs, Gottweis, Barral, Chou, Corrado, Matias, Karthikesalingam,
  and Natarajan}]{tu-2025-amie}
Tao Tu, Mike Schaekermann, Anil Palepu, Khaled Saab, Jan Freyberg, Ryutaro
  Tanno, Amy Wang, Brenna Li, Mohamed Amin, Yong Cheng, Elahe Vedadi, Nenad
  Tomasev, Shekoofeh Azizi, Karan Singhal, Le~Hou, Albert Webson, Kavita
  Kulkarni, S.~Sara Mahdavi, Christopher Semturs, and 7 others. 2025.
\newblock \href {https://doi.org/10.1038/s41586-025-08866-7} {Towards
  conversational diagnostic artificial intelligence}.
\newblock \emph{Nature}, 642:442--450.

\bibitem[{van~der Heijden et~al.(2022)van~der Heijden, Cefo, Witteman, and
  Grootens}]{vanderheijden-2022-positive-tests}
Paul~T. van~der Heijden, Irma Cefo, Cilia L.~M. Witteman, and Koen~P. Grootens.
  2022.
\newblock \href {https://doi.org/10.1016/j.comppsych.2022.152325} {On the use
  of positive test strategies when diagnosing mental disorders}.
\newblock \emph{Comprehensive Psychiatry}, 116:152325.

\bibitem[{Wang et~al.(2025{\natexlab{a}})Wang, Wang, Wu, Yang, Wang, Feng,
  Chen, Wang, and Zhang}]{wang-2025-annaagent}
Ming Wang, Peidong Wang, Lin Wu, Xiaocui Yang, Daling Wang, Shi Feng, Yuxin
  Chen, Bixuan Wang, and Yifei Zhang. 2025{\natexlab{a}}.
\newblock \href {https://doi.org/10.18653/v1/2025.findings-acl.1192}
  {{A}nna{A}gent: Dynamic evolution agent system with multi-session memory for
  realistic seeker simulation}.
\newblock In \emph{Findings of the Association for Computational Linguistics:
  ACL 2025}, pages 23221--23235, Vienna, Austria. Association for Computational
  Linguistics.

\bibitem[{Wang et~al.(2024)Wang, Milani, Chiu, Zhi, Eack, Labrum, Murphy,
  Jones, Hardy, Shen, Fang, and Chen}]{wang-2024-patient-psi}
Ruiyi Wang, Stephanie Milani, Jamie~C. Chiu, Jiayin Zhi, Shaun~M. Eack, Travis
  Labrum, Samuel~M. Murphy, Nev Jones, Kate Hardy, Hong Shen, Fei Fang, and
  Zhiyu~Zoey Chen. 2024.
\newblock \href {https://doi.org/10.18653/v1/2024.emnlp-main.711}
  {{PATIENT-$\psi$}: Using large language models to simulate patients for
  training mental health professionals}.
\newblock In \emph{Proceedings of the 2024 Conference on Empirical Methods in
  Natural Language Processing}, pages 12772--12797, Miami, Florida, USA.
  Association for Computational Linguistics.

\bibitem[{Wang et~al.(2025{\natexlab{b}})Wang, Mao, Fried, and
  Neubig}]{wang-2025-awm}
Zora~Zhiruo Wang, Jiayuan Mao, Daniel Fried, and Graham Neubig.
  2025{\natexlab{b}}.
\newblock \href {https://proceedings.mlr.press/v267/wang25bx.html} {Agent
  workflow memory}.
\newblock In \emph{Proceedings of the 42nd International Conference on Machine
  Learning}, volume 267 of \emph{Proceedings of Machine Learning Research},
  pages 63897--63911. PMLR.

\bibitem[{{World Health
  Organization}(2024{\natexlab{a}})}]{who-2024-icd11-cddr}
{World Health Organization}. 2024{\natexlab{a}}.
\newblock \href {https://www.who.int/publications/i/item/9789240077263}
  {\emph{Clinical Descriptions and Diagnostic Requirements for {ICD-11} Mental,
  Behavioural and Neurodevelopmental Disorders}}.
\newblock World Health Organization, Geneva.

\bibitem[{{World Health
  Organization}(2024{\natexlab{b}})}]{who-2024-lmm-guidance}
{World Health Organization}. 2024{\natexlab{b}}.
\newblock \href {https://iris.who.int/handle/10665/375579} {\emph{Ethics and
  Governance of Artificial Intelligence for Health: Guidance on Large
  Multi-Modal Models}}.
\newblock World Health Organization, Geneva.

\bibitem[{{World Health Organization}(2025)}]{who-2025-mental-health}
{World Health Organization}. 2025.
\newblock \href {https://www.who.int/publications/i/item/9789240113817}
  {\emph{World Mental Health Today: Latest Data}}.
\newblock World Health Organization, Geneva.

\bibitem[{{World Health Organization}(2026)}]{who-2026-icd11-mms}
{World Health Organization}. 2026.
\newblock \href {https://icd.who.int/browse/2026-01/mms/en} {{ICD-11} for
  mortality and morbidity statistics}.
\newblock Version 2026-01. Accessed October 3, 2026.

\bibitem[{Wu et~al.(2026)Wu, Wan, Li, Zhao, Ma, Ye, Zhang, Pop, Zhang, and
  Chen}]{wu-2026-wisemind}
Yuqi Wu, Guangya Wan, Jingjing Li, Shengming Zhao, Lingfeng Ma, Tianyi Ye, Mike
  Zhang, Ion Pop, Yanbo Zhang, and Jie Chen. 2026.
\newblock \href {https://doi.org/10.1038/s41746-026-02559-9} {{WiseMind}: A
  knowledge-guided multi-agent framework for accurate and empathetic
  psychiatric diagnosis}.
\newblock \emph{npj Digital Medicine}, 9:575.

\bibitem[{Xu et~al.(2026{\natexlab{a}})Xu, Ji, Lyu, Yi, Yang, Ji, Chen, Wang,
  Lan, Wang, Wang, Dong, Tian, Dong, and Hu}]{xu-2026-puppet}
Chen Xu, Yu~Ji, Zhenyu Lyu, Yang Yi, Yizhe Yang, Luyao Ji, Chaoyi Chen,
  Xianyang Wang, Tian Lan, Zhihua Wang, Juan Wang, Xunde Dong, Fuze Tian, Qunxi
  Dong, and Bin Hu. 2026{\natexlab{a}}.
\newblock \href {https://doi.org/10.18653/v1/2026.acl-long.121} {{PUPPET}:
  Neural-symbolic standardized patients for mental health}.
\newblock In \emph{Proceedings of the 64th Annual Meeting of the {A}ssociation
  for {C}omputational {L}inguistics (Volume 1: Long Papers)}, pages 2599--2634,
  San Diego, California, United States. Association for Computational
  Linguistics.

\bibitem[{Xu et~al.(2026{\natexlab{b}})Xu, Zhou, Ma, Xiao, Ding, Yan, Geng, Li,
  Han, Chen, and Deng}]{xu-2026-lingxidiagbench}
Shihao Xu, Tiancheng Zhou, Jiatong Ma, Ming Xiao, Yanli Ding, Yiming Yan,
  Haiyang Geng, Guoyi Li, Yunyun Han, Jianhua Chen, and Yafeng Deng.
  2026{\natexlab{b}}.
\newblock \href {https://doi.org/10.1145/3770855.3817539} {{LingxiDiagBench}: A
  multi-agent framework for benchmarking {LLMs} in {Chinese} psychiatric
  consultation and diagnosis}.
\newblock In \emph{Proceedings of the 32nd ACM SIGKDD Conference on Knowledge
  Discovery and Data Mining V.2}, pages 10080--10091, New York, NY, USA.
  Association for Computing Machinery.

\bibitem[{Yang et~al.(2023)Yang, Ji, Zhang, Xie, Kuang, and
  Ananiadou}]{yang-2023-interpretable}
Kailai Yang, Shaoxiong Ji, Tianlin Zhang, Qianqian Xie, Ziyan Kuang, and Sophia
  Ananiadou. 2023.
\newblock \href {https://doi.org/10.18653/v1/2023.emnlp-main.370} {Towards
  interpretable mental health analysis with large language models}.
\newblock In \emph{Proceedings of the 2023 Conference on Empirical Methods in
  Natural Language Processing}, pages 6056--6077, Singapore. Association for
  Computational Linguistics.

\bibitem[{Yao et~al.(2023)Yao, Zhao, Yu, Du, Shafran, Narasimhan, and
  Cao}]{yao-2023-react}
Shunyu Yao, Jeffrey Zhao, Dian Yu, Nan Du, Izhak Shafran, Karthik Narasimhan,
  and Yuan Cao. 2023.
\newblock \href {https://openreview.net/forum?id=WE_vluYUL-X} {{ReAct}:
  Synergizing reasoning and acting in language models}.
\newblock In \emph{The Eleventh International Conference on Learning
  Representations}.

\bibitem[{Zimmerman(2024)}]{zimmerman-2024-questionnaires}
Mark Zimmerman. 2024.
\newblock \href {https://doi.org/10.1002/wps.21191} {The value and limitations
  of self-administered questionnaires in clinical practice and epidemiological
  studies}.
\newblock \emph{World Psychiatry}, 23(2):210--212.

\end{thebibliography}

\clearpage
\appendix
\input{sections/appendix/appendix}
\AppendixEnd

\end{document}